\documentclass[conference]{IEEEtran}
\usepackage{paralist}
\usepackage{hyphenat}
\usepackage[shortcuts]{extdash}
\usepackage{graphicx}
\usepackage{subcaption}
\usepackage{pdfpages}
\usepackage{caption}
\usepackage{algorithm}
\usepackage{algpseudocode}
\usepackage{booktabs}
\usepackage{commath}
\usepackage{amsthm}
\usepackage{tikz}
\usepackage{cite}
\usepackage{amsmath,amssymb,amsfonts}
\usepackage{textcomp}
\usepackage{xcolor}
\usepackage{multirow}
\usepackage{hyperref}

\def\BibTex{{\rm B\kern-.05em{\sc i\kern-.025em b}\kern-.08em h T\kern-.1667\lower.7ex\hbox[E]\kern-.125emX}}

\algnewcommand\algorithmicforeach{\textbf{for each}}
\algdef{S}[FOR]{ForEach}[1]{\algorithmicforeach\ #1\ \algorithmicdo}

\usepackage{graphicx}

\AtBeginDocument{%
  \providecommand\BibTeX{{%
    \normalfont B\kern-0.5em{\scshape i\kern-0.25em b}\kern-0.8em\TeX}}}

\author{
\IEEEauthorblockN{Seyedeh Gol Ara Ghoreishi}
\IEEEauthorblockA{
\textit{Florida Atlantic University} \\
Boca Raton, USA \\
sghoreishi2021@fau.edu
}
\and
\IEEEauthorblockN{Sonia Moshfeghi}
\IEEEauthorblockA{
\textit{Florida Atlantic University} \\
Boca Raton, USA \\
smoshfeghi2021@fau.edu
}
\and
\IEEEauthorblockN{Muhammad Tanveer Jan}
\IEEEauthorblockA{
\textit{Florida Atlantic University} \\
Boca Raton, USA \\
mjan2021@fau.edu
}

\and
\IEEEauthorblockN{Joshua Conniff}
\IEEEauthorblockA{
\textit{Florida Atlantic University} \\
Boca Raton, USA \\
fau.jconniff@health.fau.edu
}

\and
\IEEEauthorblockN{KwangSoo Yang}
\IEEEauthorblockA{
\textit{Florida Atlantic University} \\
Boca Raton, USA \\
yangk@fau.edu
}

\and
\IEEEauthorblockN{Jinwoo Jang}
\IEEEauthorblockA{
\textit{Florida Atlantic University} \\
Boca Raton, USA \\
jangj@fau.edu
}
\and
\IEEEauthorblockN{Borko Furht}
\IEEEauthorblockA{
\textit{Florida Atlantic University} \\
Boca Raton, USA \\
bfurht@fau.edu
}
\and
\IEEEauthorblockN{Ruth Tappen}
\IEEEauthorblockA{
\textit{Florida Atlantic University} \\
Boca Raton, USA \\
rtappen@health.fau.edu
}
\and
\IEEEauthorblockN{David Newman}
\IEEEauthorblockA{
\textit{Florida Atlantic University} \\
Boca Raton, USA \\
dnewma14@fau.edu
}
\and
\IEEEauthorblockN{Monica Rosselli}
\IEEEauthorblockA{
\textit{Florida Atlantic University} \\
Boca Raton, USA \\
mrossell@fau.edu
}

\and
\IEEEauthorblockN{Jiannan Zhai}
\IEEEauthorblockA{
\textit{Florida Atlantic University} \\
Boca Raton, USA \\
jzhai@fau.edu
}

}

\begin{document}

% \IEEEoverridecommandlockouts
% \IEEEpubid{\makebox[\columnwidth]{978-1-5386-5541-2/18/\$31.00~\copyright2018 IEEE \hfill}
% \hspace{\columnsep}\makebox[\columnwidth]{ }}
\title{Anomalous Behavior Detection in Trajectory Data of Older Drivers}
\maketitle

\newpage

% Copyright notice in footnote
\renewcommand{\thefootnote}{\fnsymbol{footnote}}
\setcounter{footnote}{0}
\footnotetext{%
  Copyright © 2023 IEEE. Personal use of this material is permitted. Permission from IEEE must be obtained for all other uses, in any current or future media, including reprinting/republishing this material for advertising or promotional purposes, creating new collective works, for resale or redistribution to servers or lists, or reuse of any copyrighted component of this work in other works.
}

\begin{abstract}
Given a road network and a set of trajectory data, the anomalous behavior detection (ABD) problem is to identify drivers that show significant directional deviations, hard-brakings, and accelerations in their trips. The ABD problem is important in many societal applications, including Mild Cognitive Impairment (MCI) detection and safe route recommendations for older drivers. The ABD problem is computationally challenging due to the large size of temporally-detailed trajectories dataset. In this paper, we propose an Edge-Attributed Matrix that can represent the key properties of temporally-detailed trajectory datasets and identify abnormal driving behaviors. Experiments using real-world datasets demonstrated that our approach identifies abnormal driving behaviors.
\end{abstract}
                              
\section{Introduction}

\begin{figure*}[t!]
    \centering
    \begin{subfigure}[t]{0.55\textwidth}
	\includegraphics[width=8cm]{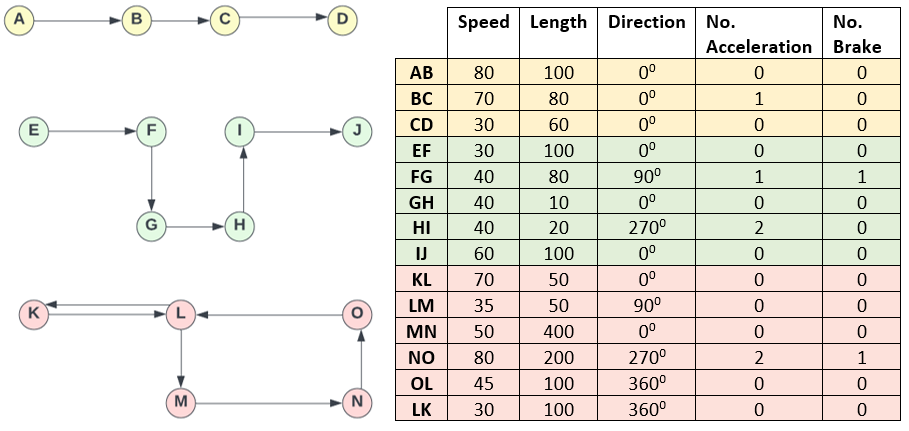} 
	\caption{The input matrix and the graphs.}
	\label{fig:input-graph}
    \end{subfigure}	
    \begin{subfigure}[t]{0.40\textwidth}
	\includegraphics[width=5.5cm]{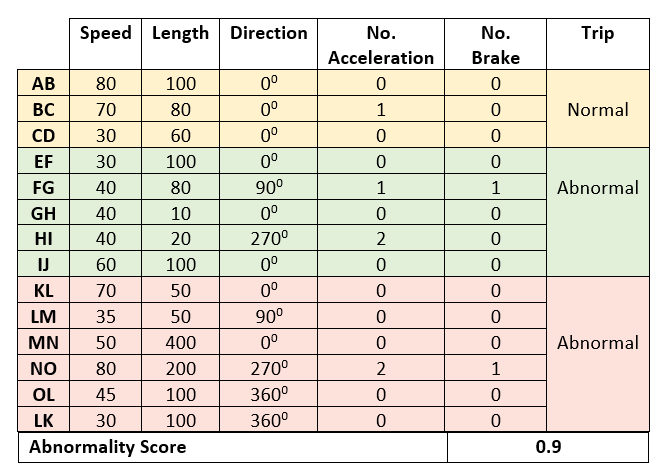} 
	\caption{The output matrix.}
	\label{fig:output}
    \end{subfigure}
	\caption{An example of the input and output of ABD -three trips with its corresponding input and output matrices.}
	\label{fig:example}
\end{figure*}

Given a road network and a set of trajectory data, the anomalous behavior detection (ABD) problem is to identify drivers that show significant directional deviations, hard-brakings, and accelerations in their trips. Consider an example input of ABD in Figure~\ref{fig:input-graph}. Each graph represents a trip on the road network and the table shows the temporally detailed driving information (i.e., speed, length, direction and the number of accelerations and brakes) for edges. Figure \ref{fig:output} shows an example output of ABD containing the abnormality score for each driver. One of the most appealing properties of our model is the ability to capture geometrical (e.g., longitude and latitude), directional, and topological features of the trajectories. Analyzing this data presents a considerable challenge due to the large size of temporally-detailed trajectory datasets. 

\subsection{Application Domain}
ABD problem could be linked to identifying cognitive impairment in the elderly drivers\footnote{\url{https://www.fau.edu/newsdesk/articles/older-drivers-research.php}}.
Older drivers face many challenges during their trips. They might miss intended routes or exits, make incorrect turns, and go the wrong way due to reduced spatial awareness or the need to remember the way when trying to reach their destination. Some trips had cyclic patterns and deviated significantly from directed paths. However, not all deviations are necessarily abnormal. It is possible for drivers to make turns and take alternative routes due to various reasons such as roadblocks, traffic congestion, or emergencies \cite{b1}. Furthermore, some drivers might choose the correct routes during their journeys, but they exhibit frequent sudden accelerations. These abrupt speed changes could also signal confusion or a lack of awareness regarding the road network conditions.

\subsection{Problem Definition}
In our problem formulation, a trip is represented as a directed graph consisting of nodes, edges and edge attributes. Each node represents a spatial location on the road network and each edge represents a road segment. Every edge has temporally-detailed driving attributes. The ABD problem can be formalized as follows:\\
\\

\noindent\textbf{Input:}
\begin{compactitem}
\item A road network consisting of a set of nodes $V$ and a set of edges $E$,
\item a set of trip attributes on a road segment $A: E \rightarrow R_{0}^+$, and
\item the minimum length of a trip $\alpha$
\end{compactitem}
\noindent
\textbf{Output:} Abnormal driving behaviors\\
\textbf{Objective:}
\begin{compactitem}
\item Compute the abnormality scores related to direction deviations, hard brakings, and hard accelerations. 
\end{compactitem}
\textbf{Constraints:}
\begin{compactitem}
\item Every length of the trip should be greater than $\alpha$.
\end{compactitem}

\subsection{Outline}
The next of this paper is organized as follows:
Section 2 describes the related work. Section 3 introduces the proposed method. Section 4 illustrates the experimental dataset and results. Section 5 discusses future directions and concludes the research work.
\section{RELATED WORK}
Numerous researchers have analyzed the GPS trajectories of vehicles in a road network and calculated the similarity between different trajectories. However, most of their attention has been on vehicles traveling from one specific starting point to a particular destination. Researchers apply various methods to identify abnormal behaviors in these driving patterns. In paper \cite{b2}, GPS points are aggregated into spatiotemporal units, where each unit represents the maximum value of speed, acceleration, and direction deviation observed within a specific road segment. The authors compared these units among drivers to identify those showing significant deviations from typical traffic patterns and proposed a variational autoencoder method named STDTB-AD to quantify the level of abnormality for each driver. The researchers in \cite{b3} used trajectory data to predict the traffic context on road networks and created a model to represent road segments and capture characteristics of road networks. Another paper, \cite{b4} proposed a partition and detect approach to achieve anomalous sub-trajectory identification. The process begins by splitting each trajectory into t-partitions. Then, it detects segments that significantly differ from those in other trajectories in terms of distance and shape. A trajectory with significant abnormal segments contains an outlier sub-trajectory and is classified as an outlier. In \cite{b5}, a self-supervised proposed approach creates low-dimensional vectors from raw trajectories. The method first presents raw trajectory data and road networks as road segment vectors and then combines these vectors into a single representation called trajectory vectors. This transformation is achieved by maintaining spatial-temporal characteristics of the trajectories. In \cite{b6}, each trajectory is segmented to a minimum number of homogeneous segments with similar spatiotemporal characteristics, such as location, heading, speed, velocity, sinuosity, curviness, and shape. Authors in \cite{b7} clustered trajectories by using the iVAT algorithm and employing a two-stage process. In the first stage, a similarity measure is applied to group trajectories that follow similar paths but may have opposite starting and ending points. These similar trajectories are assigned to the same cluster. In the second stage, directional similarity is used within each cluster to separate trajectories going in opposite directions. From these clustered trajectories, the proposed approach identify trajectory outliers by finding trajectories significantly distant from others in the same cluster or by detecting clusters with too few trajectories. This process helps identify unusual or isolated trajectory patterns within the dataset. In \cite{b8}, authors proposed a distance-based method to establish local clusters for continuous trajectory streams and use pruning techniques to monitor anomalous behavior. In \cite{b9}, the main idea is to detect fraud in taxi driving by combining density and distance characteristics. The similarity is that first, they compute the expected distance of the most common routes and the distances a trajectory differs from the norm.
In Graph-based methods for identifying outliers in trajectories, in \cite{b10}, the researchers depict various trajectory trips but only for trips with the same starting and ending points. They analyzed the common nodes and edges between these graphs to detect trajectory outliers. If a graph has edges that are notably different from the usual patterns observed in other graphs or contains a different subgraph between the source and target graph, it is considered as an outlier. A machine learning algorithm employs the extracted features to classify the trajectory. In \cite{b11}, researchers detected newly generated cycles in a dynamic graph that is constantly changing. Regarding detour detection, in \cite{b12} for the same start and end point, a detour is defined as taking much time or driving long distances. The authors in \cite{b13} proposed a graph-based method to detect loop closure for localization and mapping. For abnormal taxi trajectory, authors of \cite{b14} used detour detection, local shape and speed anomaly detection by using spatial location, sequence, and behavioral features in vehicle trajectories. In their approach, detour detection is based on how much a trajectory changes direction at a specific point in its path. In\cite{b15}, a distance metric is proposed to measure the similarity between anomalous and normal trajectories in different driving patterns, such as long-distance detours. They establish definitions for global detour and local detour. A global detour refers to a considerably longer trajectory that significantly deviates from the norm, while a local detour is a longer trajectory with partial deviations from the norm. Anomalous behavior like detours in  \cite{b1} are based on the density of heading changes and the combination of sequential turns in the same direction. Their method automatically identifies a detour's start and end points and computes the detour factor while retaining the dynamic aspects of trajectory data. To achieve this, they employ a recursive Bayesian filter to estimate a single unobserved state and make predictions.

\section{METHODOLOGY}
We propose a framework to identify unusual driving patterns or behaviors that deviate from the expected norms during a trip. The primary objective is to evaluate sudden speed changes (like rapid slowing down or speeding up)  and the paths a driver chooses in a trip. This evaluation relies on individual trajectory data points and subsequently assigns a numerical score reflecting the abnormality of the trip.  In our study, a trip characterized by a long-distance cyclic pattern or a looping behavior should receive a high abnormality score. However, it is evident that all deviations, such as roundabouts or short-distance U-turns, are not abnormal or unusual behavior. As well as, many driving scenarios require drivers to make long detours or take alternative routes for various reasons, such as roadblocks or obstacles, traffic congestion, or emergencies\cite{b1}. In these cases, in our framework, apart from the length of the deviation paths, we also need to consider the speed changes during these detours. This is because an increase in speed during deviation paths suggests that the drivers have an acceptable reason. Besides, deviations allow them to drive at higher speeds and reach their destination more quickly. Our method includes data preprocessing, spatial graph modeling, and anomaly detection. 

\begin{figure}[h!]
    \centering
    \begin{subfigure}[t]{0.45\textwidth}
	\includegraphics[width=3.0in]{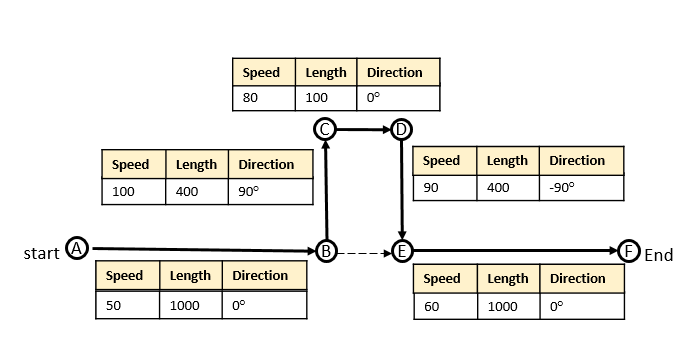} 
	\caption{Edge-Attributed Graph}
	\label{fig:step1-graph}
    \end{subfigure}	
    \begin{subfigure}[b]{1.5in}
	\includegraphics[width=1.4in]{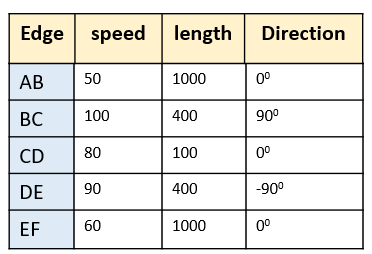}   
	\caption{Edge-Attributed Matrix}
	\label{fig:step1-table}
	\end{subfigure}	

	\caption{Example of a trip with a set of trajectory attributes}
	\label{fig:steps}
\end{figure}

\subsection{Data preprocessing}
Since our main goal is to concentrate on the drivers' paths, it is necessary to preprocess temporally-detailed trajectory data within the road network. The temporally-detailed trajectory data contains GPS points, and we project them onto their nearest road segments to identify which points are located on which roads. This process enables us to explain the movement patterns of a trajectory throughout time and space. To model the trajectory data and their topological connectivity, we create a spatial graph for each driver’s trip. Fig. 2a shows an edge-attributed graph consisting of 6 nodes and 5 edges, and each edge is associated with 3 attributes that provide essential indications about the associated road segments. The features include the average direction and speed of trajectory data and the segment's length. To simplify the example, only three attributes are selected for this illustration. These attributes are further presented in matrix format in Fig. 2b.

•\quad\textbf{Definition 1} (Trajectory data points). In trajectory data, each point $\mathrm{p}_\mathrm{i}$ represents the trajectory's state at a particular time. This point is characterized by properties, including location, speed, direction and occurrence of hard acceleration and hard brake which is shown as:

\begin{equation}
\mathrm{p}_{\mathrm{i}}^{\mathrm{t}}=\left\{\operatorname{lat}_{\mathrm{i}}^{\mathrm{t}}, 
\operatorname{lon}_{\mathrm{i}}^{\mathrm{t}},
\mathrm{v}_{\mathrm{i}}^{\mathrm{t}},
\mathrm{d}_{\mathrm{i}}^{\mathrm{t}}, \mathrm{acc}_{\mathrm{i}}^{\mathrm{t}},
\mathrm{br}_{\mathrm{i}}^{\mathrm{t}}\right\},
\end{equation}
 
 where $\mathrm{lat}$ is latitude, $\mathrm{lon}$ is longitude, $\mathrm{v}$ is velocity, $\mathrm{d}$ is direction, $\mathrm{acc}$ is acceleration, $\mathrm{br}$ is brake, superscript $\mathrm{t}$ represents the time t, and subscript $\mathrm{i}$ represents the id of the point.
 
•\quad\textbf{Definition 2}  (Edge-Attributed Matrix). Every trip is represented by a directed graph. Each edge has a set of attributes that are listed in a matrix.

•\quad\textbf{Definition 3} (Edge attributes). The edge attributes are the average speed and direction, length and the total number of hard brakes and hard accelerations observed on its corresponding road segment over time.

\subsection{Spatial graph model}
With the Edge-Attributed Matrix, we can examine a variety of spatial graphs with various source and destination points. For a given set of graphs, we compute the anomaly scores using the isolation Forest algorithm to determine which graphs are anomalous and differ from the others in terms of attributes. 
\begin{figure*}[!htbp]
    \centering
    \includegraphics[width=\textwidth]{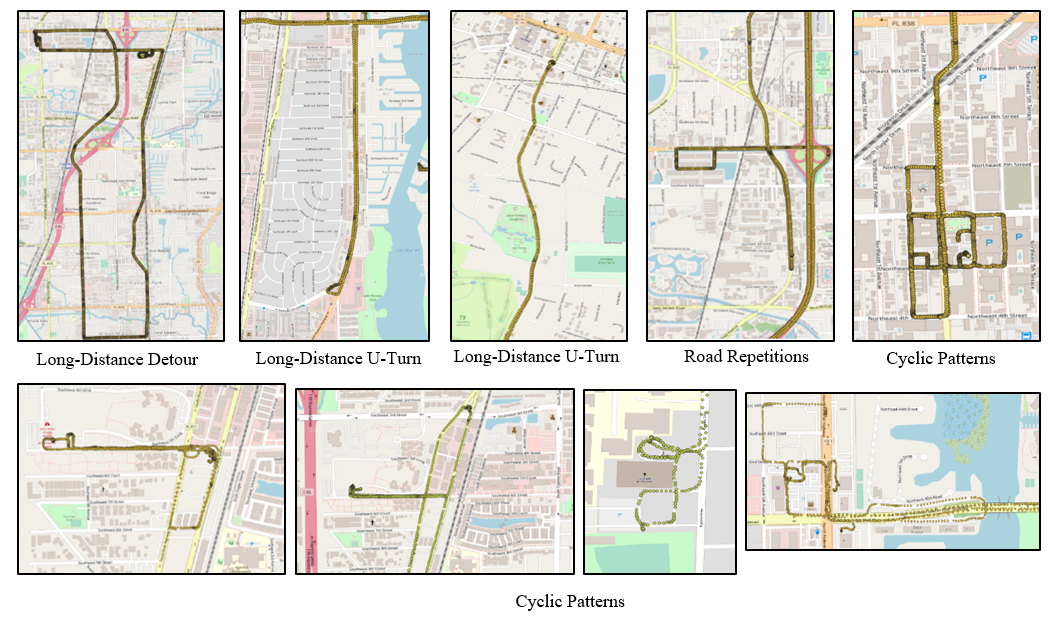} 
    \caption{Snapshot of the Real Dataset - A visual representation of the dataset used in the study, with each picture representing a random trip}
\end{figure*}

\subsection{Anomaly detection using Isolation Forest (iForest)}
Isolation Forest is an unsupervised anomaly detection algorithm to find rare and distinct data points within a dataset. It achieves this by constructing multiple binary trees\cite{b17}. These trees randomly select features and values to divide the data into smaller groups through a recursive process\cite{b17}. Anomalies often require fewer splits to be separated in these trees because they are distinct from regular data. In this algorithm, the contamination parameter serves as a threshold for defining outliers in the dataset. Isolation Forest is useful for finding outliers and anomalies in various datasets, including those with unbalanced anomaly distributions\cite{b17}.

\begin{figure*}[!htbp]
	\centering
	 \hspace{0.0in}  
		\includegraphics[width=5in]{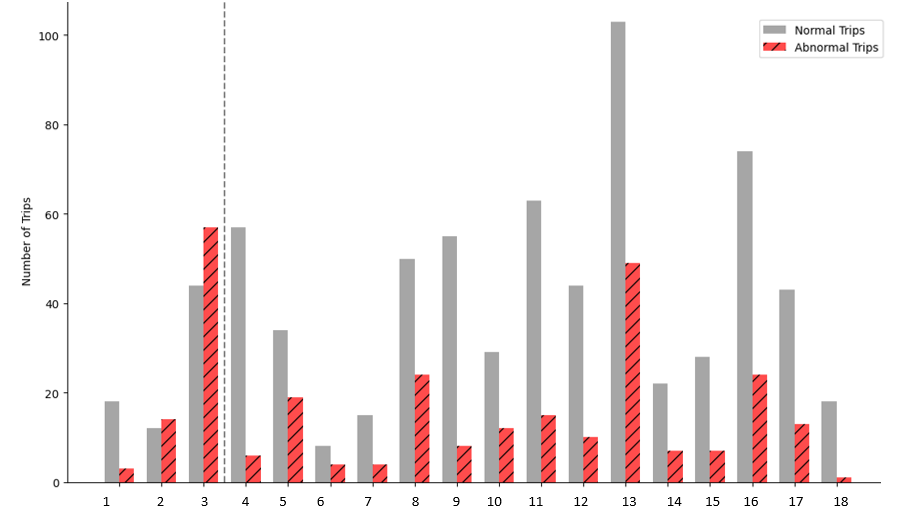} 
		\caption{Number of normal and abnormal trips for each sample}
		\label{fig:step1-graph}	
\end{figure*}

\section{EXPERIMENTS}
In this section, we describe the generated data and display the findings with the iForest algorithm.

\subsection{Data}
Our 5-year research project funded by the National Institutes of Health (NIH) comprises drivers with an age group between 65 and 85 years old\cite{b16}. In our dataset, the participant's behavior is monitored over 3 years. We collect 18 senior citizens' trajectory data between May 1, 2022 and July 30, 2022.
The data is recorded with in-vehicle telematics sensors. The telematics sensor provides Inertial Measurement Unit (IMU) and GPS data. IMU data records the vehicle's dynamic motions and orientations. It consists of two components:

\begin{compactitem}
\item 3-Axial accelerometer: This device records how the vehicle accelerates or decelerates in three directions (X, Y, Z). It detects events like hard braking (sudden stops) or hard accelerations (rapid speed increases) \footnote{\url{https://www.autopi.io/glossary/accelerometer/}\label{myfootnote}}.
\item 3-Axial Gyroscope (Angular Velocity): This measures the rate at which the vehicle is rotating or changes in direction\footnotemark[2]. Fig.4 shows the Accelerometer and Gyroscope\footnotemark[2].
\end{compactitem}
\begin{figure}[h!]
    \centering
	\includegraphics[width=3.0in]{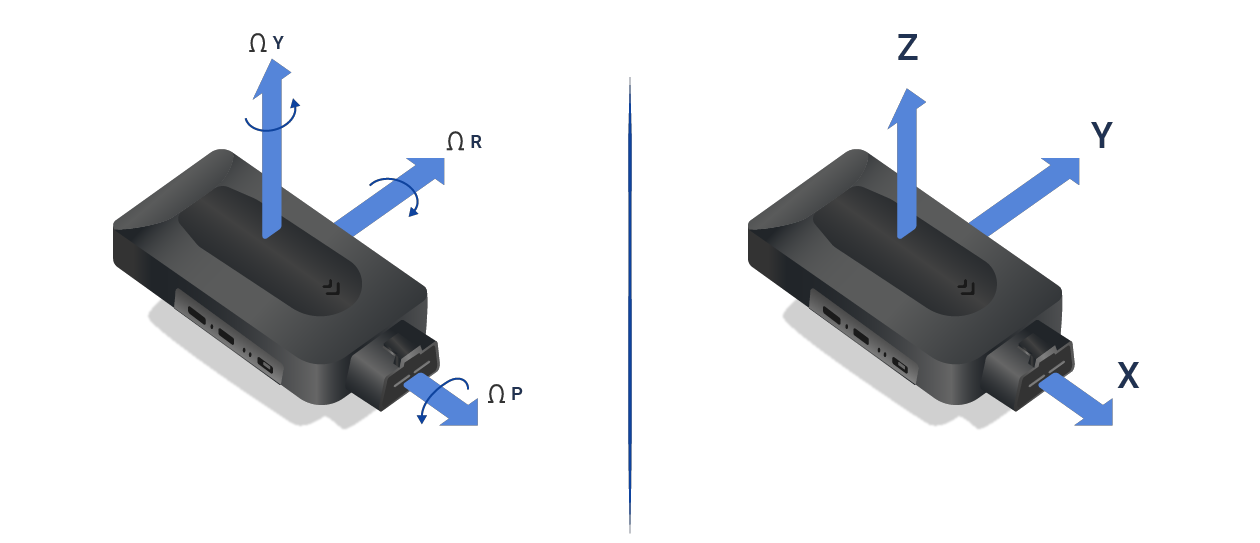} 
	\caption{Gyroscope sensing angular orientation and Accelerometer sensing Axis orientation \protect\footnotemark[2] }
	\label{fig:step2-graph}
\end{figure}
GPS Data: GPS data provides information about the vehicle's location and movement in the global positioning system. It includes several key components: timestamp, latitude, longitude, and altitude to analyze travel distance and Course Over Ground (COG) to calculate the heading or direction in which the vehicle is moving.

In a road network, the trajectories exhibit various driving patterns, and Fig.3 illustrates several snapshots that correspond to different drivers, each depicting potential abnormal driving behaviors among the elderly. These examples capture situations such as repetition on the road, lengthy detours, U-turns, cyclic patterns, and instances where drivers seem to have lost their way. Road segments comprise their trajectory points across time and by using the raw dataset, we compute statistics. The average velocity and COG, the sum of hard brakes and hard accelerations of data points with road segment length were used as input data. We apply this processing method and iForest to the dataset and obtain a result at different spatial scales.	

\section{Results}
The iForest algorithm calculates trip anomaly scores ranging from 0 to 1. Values close to 1 are considered as anomalous. In this study, the threshold to classify trips is set to 0.6. Our approach extends to driver classification, where the mean anomaly score for all trips informs whether a driver is classified as normal or abnormal. We define the threshold for drivers based on the top 10\%  and 20\% \cite{b16} of the total drivers' anomaly scores. The contamination parameter, which indicates the likelihood of outliers in the dataset, is set to 0.2. Fig.5 presents the count of normal and abnormal trips for each driver. The vertical line indicates the top 20\%  threshold within the drivers' anomaly score distribution and distinguishes the 3 abnormal drivers.

The dataset employed in our experiments contains labels only for drivers, and there are no labels for individual trips. Besides, this dataset is unique and has not been previously examined in the literature. Consequently, making direct comparisons with existing studies is challenging due to this disparity in labeling.
To evaluate the effectiveness of our approach in detecting abnormal drivers, we employed evaluation metrics of accuracy(ACC) and F1-score. These metrics encompass True Positive (TP), True Negative (TN), False Positive (FP), and False Negative (FN) values. ACC quantifies the percentage of correctly classified cases throughout the entire dataset, precision (P) assesses the proportion of true positives among all predicted positives and recall (R) measures the proportion of true positives among actual positives. It was shown in the literature that F1-score is more suitable than the 
accuracy\cite{b18}. 

\begin{equation}
A C C=\frac{T P+T N}{T P+T N+F P+F N}
\end{equation}

\begin{equation}
P=\frac{T P}{T P+F P}
\end{equation}

\begin{equation}
R=\frac{T P}{T P+F N}
\end{equation}

\begin{equation}
F 1=\frac{2 * P * R}{P+R}
\end{equation}

Table 1 illustrates the relationship between the chosen thresholds and performance metrics. The accuracy is notably higher when the top 10\% of drivers are considered. However, for the top 20\%, the improvement in F1 score suggests that selecting a broader group of drivers results in a better balance between precision and recall.

\begin{table}
    \centering
    \caption{Anomaly detection metrics}
    \begin{tabular}{cccccc}
        \toprule
        Contamination & & Top10\%&Top20\%\\

        \midrule
        \multirow{2}{*}{0.2} & ACC &0.83 &0.77\\
        & F1 & 0.40 & 0.50\\   
        \bottomrule
    \end{tabular}
\end{table}

\section{CONCLUSION AND FUTURE WORK}
We discussed the problem of automatically identifying abnormal driving habits among senior individuals. The problem is challenging because of the large size of the road network and trajectory data. By utilizing in-vehicle sensor data, edge-attributed matrices, and isolation forest algorithms, we aim to provide a baseline approach for identifying unusual actions, such as cyclic routes and long-distance detours, and high acceleration or braking incidents, which may indicate cognitive decline or impairments in older drivers. However, there is still room for improvement and additional study in this field.\\
In future work, we plan to expand our dataset by incorporating more samples of subjects for the test. Additionally, we intend to design and evaluate a novel algorithm for the notable characteristics of senior drivers.

\section{Acknowledgments}
This material is based upon work supported by the National Science Foundation CAREER under Grant No. 1844565. and the National Institutes of Health under Grant No. 1R01AG068472.

\end{document}